\documentclass[11pt,a4paper]{article}
\pdfoutput=1
\usepackage{authblk}
\usepackage[hyperref]{emnlp2020}
\usepackage{graphicx}
\usepackage{amsmath}
\usepackage{float}
\usepackage{enumitem} % Allows use of [nosep] for lists
\usepackage{forest} % Draw figure 1
\usepackage{times}
\usepackage{latexsym}

\newcommand\blfootnote[1]{%
  \begingroup
  \renewcommand\thefootnote{}\footnote{#1}%
  \addtocounter{footnote}{-1}%
  \endgroup
}

\usepackage{microtype}

\aclfinalcopy % Uncomment this line for the final submission
\def\aclpaperid{1569} %  Enter the acl Paper ID here

\title{Exploring Contextualized Neural Language Models for Temporal Dependency Parsing}

\author[1]{Hayley Ross$\dagger$}
\author[2]{Jonathon Cai}
\author[2]{Bonan Min}
\affil[1]{Harvard University}
\affil[ ]{\texttt{hayleyross@g.harvard.edu}}
\affil[2]{Raytheon BBN Technologies}
\affil[ ]{\texttt { \{jonathon.cai, bonan.min\}@raytheon.com}}

\date{}

\begin{document}
\maketitle
\begin{abstract}
Extracting temporal relations between events and 
time\blfootnote{A previous version is available at \url{https://arxiv.org/abs/2004.14577}.}
expressions\blfootnote{$\dagger$ Work done during an internship at BBN.} 
has many applications such as constructing event timelines and time-related question answering. It is a challenging problem which requires syntactic and semantic information at sentence or discourse levels, which may be captured by deep contextualized language models (LMs) such as BERT \citep{devlin2019bert}. 
In this paper, we develop several variants of BERT-based temporal dependency parser, and show that BERT significantly improves temporal dependency parsing \citep{zhang2018neural}. We also present a detailed analysis on why deep contextualized neural LMs help and where they may fall short. Source code and resources are made available at \url{https://github.com/bnmin/tdp_ranking}. 
\end{abstract}

\section{Introduction} \label{section:intro}

Temporal relation extraction has many applications such as constructing event timelines for news articles or narratives as well as time-related question answering. Recently, \citet{zhang2018structured} presented Temporal Dependency Parsing (TDP), which organizes time expressions and events in a document to form a Temporal Dependency Tree (TDT). Given a previous step which detects time expressions and events, TDP extracts the temporal structure between them.
%(More details can be found in Section~\ref{section:related}.). 
Consider this example:

{\bf Example 1:} {\it Kuchma and Yeltsin \underline{signed} a cooperation plan on \underline{February 27, 1998}. Russia and Ukraine \underline{share} similar cultures, and Ukraine was \underline{ruled} from Moscow for centuries. Yeltsin and Kuchma \underline{called} for the ratification of the treaty, \underline{saying} it would \underline{create} a ``strong legal foundation''.}

\begin{figure}[t]
\begin{center}
\begin{forest}
for tree={
    ellipse,
    draw,
    semithick,
    font=\footnotesize,
    inner sep=1.4pt, % size of ellipses
    s sep=1.1em, % horizontal spacing
    l sep=0em, % vertical spacing
    l=2.2em, % child positioning
}
[root, draw=black!60!green
    [DCT, draw=black!60!green, edge label={node[yshift=2pt,midway,left,font=\scriptsize] {depends on}} 
        [share, draw=blue, edge label={node[yshift=2pt,midway,left,font=\scriptsize] {overlap}}]
        [ruled, draw=blue, edge label={node[xshift=-2pt,midway,right,font=\scriptsize] {before}}]
        [called, draw=blue, edge label={node[yshift=2pt,midway,right,font=\scriptsize] {before}}
            [saying, draw=blue, edge label={node[midway,left,font=\scriptsize] {overlap}}
                [create, draw=blue, edge label={node[midway,left,font=\scriptsize] {after}}]
            ]
        ]
    ]
    [{Feb 27, 1998}, draw=black!60!green, edge label={node[yshift=2pt,midway,right,font=\scriptsize] {depends on}}
        [signed, draw=blue, edge label={node[midway,left,font=\scriptsize] {overlap}}]
    ]
]
\node (B1) at (current bounding box.east) [right=1ex, above=1.1em, black!60!green]
{$\left.\rule{0em}{2.2em}\right]$
};
\node (B2) at (B1.south) [below=-0.5em, blue]
{$\left.\rule{0em}{3.35em}\right]$
};
\node at (B1.east) [right=-1.5ex, below=0em, rotate=90, black!60!green, font=\scriptsize] {TIMEX};
\node at (B2.east) [right=-1.5ex, below=0em, rotate=90, blue, font=\scriptsize] {EVENTS};
\end{forest}
\end{center}
\vspace{-0.5cm}
\caption{Temporal Dependency Tree of Example 1. DCT is Document Creation Time (March 1, 1998)}
\label{fig:parsetree}
\vspace{-0.7cm}
\end{figure}
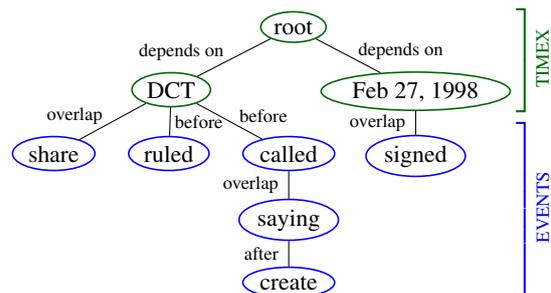

Figure \ref{fig:parsetree} shows the corresponding TDT. Compared
to previous pairwise approaches for temporal relation extraction based on TimeML \citep{pustejovsky2003timeml} which require $\binom{n}{2}$ pairs of temporal relations to be annotated, TDT significantly reduces the annotation complexity while still preserving the essential temporal structure between events and temporal relations. TDP is still challenging because it requires syntactic and semantic information at sentence and discourse levels.

Recently, deep language models such as BERT \citep{devlin2019bert} have been shown to be successful at many NLP tasks, because (1) they provide contextualized
word embeddings that are pre-trained with very large corpora, and (2) BERT in particular is shown to capture
syntactic and semantic information (\citealp{tenney2019bert}, \citealp{clark2019does}), which may include but is
not limited to tense and temporal connectives. Such information is relevant for TDP.

In this paper, we present BERT-based TDP models, and empirical evidence demonstrating that BERT significantly improves TDP. We summarize the
contributions of this paper as follows:
\begin{itemize}[nosep]
    \item We develop temporal dependency parsers that incorporate BERT, from
a straightforward usage of pre-trained BERT word embeddings, to using BERT's multi-layer multi-head self-attention architecture as an encoder  trained within an end-to-end system.
\newpage
\item We present experiments showing significant advantages of the BERT-based TDP models. Experiments show that BERT improves TDP performance in all models, with the best model achieving a 13 absolute F1 point improvement over our re-implementation of the neural model in \citep{zhang2018neural}\footnote{We were unable to replicate the F1-score reported for this corpus in \citet{zhang2019acquiring}. The improvement over the reported, state-of-the-art result is 8 absolute F1 points.}. 
\item We lay out a detailed analysis on BERT's strengths and limitations for this task.
\end{itemize}

We present technical details, experiments, and analysis in the rest of this paper.

\section{Related Work}\label{section:related}

Much previous work has been devoted to classification and annotation of relations between events
and time expressions, notably 
TimeML \citep{pustejovsky2003timeml} and TimeBank \citep{pustejovsky2003timebank}, as well as many extensions of it (see \citealp{derczynski2017automatically} for an overview). TimeML annotates all explicit relations in the text; at the extreme,
TimeBank-Dense \citep{cassidy2014timebankdense}  annotates all $\binom{n}{2}$ pairs of relations. Pair-wise annotation has three problems: $O(n^2)$ complexity; the possibility of inconsistent predictions such as $A$ before $B$, $B$ before $C$, $C$ before $A$; and forcing annotation of relations left unclear by the document. 

While extracting time expressions and events is well handled (e.g.~\citealp{strotgen2010heideltime}, \citealp{lee2014context}), relating them is still a challenging task.
Previous research on extracting these relations (e.g.~\citealp{bethard2017semeval}, \citealp{ning2017structured}, \citealp{lin2019bert}) almost always uses pair-wise TimeML-annotated data which has rich annotation but also inherits the above three complexity and consistency issues.
To address these issues, \citet{zhang2018structured} present
a tree structure of relations between time expressions and events (TDT), along with a BiLSTM model \citep{zhang2018neural} for parsing text into TDT and a crowd-sourced corpus \citep{zhang2019acquiring}. 

Organizing time expressions and events into a tree has a number of advantages over traditional pair-wise temporal annotation. It reduces the annotation complexity to $O(n)$ and avoids cyclic inconsistencies both in the annotation and the model output. Despite the apparent reduction in labeled edges, many additional edge labels can be deduced from the tree: in Figure 1, we can deduce e.g. that \emph{ruled} is before \emph{share} because \emph{ruled} is before DCT but \emph{share} overlaps DCT. A final advantage of TDTs is that they allow underspecification where the source document does not explicitly specify an order, such as the relation between \emph{signed} and \emph{called} (likely to be \emph{overlap}, but it is not certain). 
\citet{zhang2019acquiring} is currently the only English-language TDP corpus, comprising 196 newswire articles.

In addition, this paper capitalizes on the now well-documented recent advances provided by BERT \citep{devlin2019bert}. Besides offering richer contextual information, BERT in particular is shown to capture syntactic and semantic properties (\citealp{tenney2019bert}, \citealp{clark2019does}) relevant to TDP, which we show yield improvements over \citeauthor{zhang2018neural}'s original model.

\section{BERT-based Neural Models for Temporal Dependency Parsing}\label{section:approach}

Following \citet{zhang2018neural}, we transformed temporal dependency parsing (TDP) to a ranking problem: given a child mention (event or time expression) $x_i$ extracted by a previous system, the problem is to select the most appropriate parent mention from among the root node, DCT or an event or time expression from
the window $x_{i-k}, \ldots, x_i, \ldots, x_{i+m}$~\footnote{We set $k=10, m=3$ in all experiments. } around $x_i$, along with the relation label ({\it before}, {\it after}, {\it overlap}, {\it depends on}). That is, for each $x_j$ in the window, the model judges the child-parent candidate pair $\langle x_i, x_j \rangle$.
A Temporal Dependency Tree (TDT) is assembled with an incremental algorithm which selects, for each event and time expression in sequence in the document,
the highest-ranked prediction $\langle$parent, relation type$\rangle$. The tree structure is enforced by selecting the highest probability parent which does not introduce a cycle\footnote{In practice, this step to avoid cyclic edges is rare: it is required for less than $4\%$ of the predicted edges.}.

\begin{figure}[t]
\begin{center}
\scalebox{1.0}{
\includegraphics[width=1.0\linewidth]{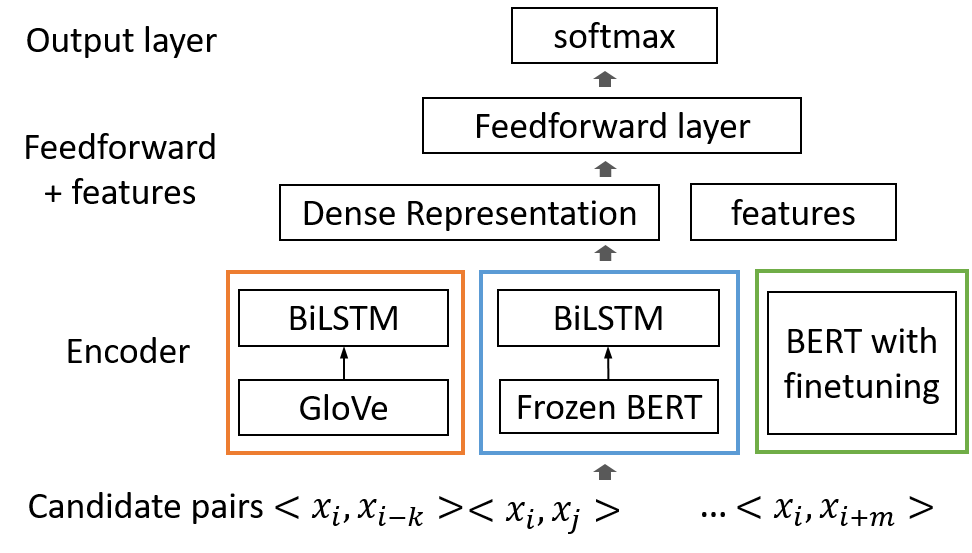}}
\end{center}
\vspace{-0.5cm}
\caption{Model architecture for TDP with three different encoders (orange, blue, green boxes). Shown with the $\langle$parent, child$\rangle$ input pairs for a given child (event or time expression) $x_i$. For simplicity, we did not show $\langle x_i, root \rangle$ and $\langle x_i, DCT \rangle$, which are included as candidate pairs for all $x_i$. 
}
\label{fig:arch}
\vspace{-0.4cm}
\end{figure}

We developed three models that share a similar overall architecture (Figure \ref{fig:arch}):
the model takes a pair of mentions (child and potential parent) as input and passes each
pair through an encoder which embeds the nodes and surrounding context into a dense
representation. All models use the same window approach described above to source parent candidates. Following \citet{zhang2018neural}, linguistic features are concatenated onto the dense representation,
which is then passed to a feed-forward layer and a softmax function to generate scores
for each relation label for each pair.

We developed three types of encoder:
%\vspace{-0.1cm}
\begin{itemize}[nosep]
  \item \textsc{bilstm} and \textsc{bilstm-glove} feed the document's word embeddings to a BiLSTM to encode the pair as well as the surrounding context. The word embeddings can be either randomly initialized (identical to \citealp{zhang2018neural}) (in \textsc{bilstm}), or pre-trained from a large corpus -- we used GloVe \citep{pennington2014glove} (in \textsc{bilstm-glove}).
%  \vspace{-0.1cm}
    \item \textsc{bilstm-bert} replaces the word embeddings with frozen (pre-trained) BERT contextualized word embeddings. We used the BERT-base uncased model\footnote{\url{https://github.com/google-research/bert}}, which has been trained on English Wikipedia and the BookCorpus.
%    \vspace{-0.1cm}
  \item \textsc{bert-ft}: BERT's multi-layer multi-head self-attention architecture (with pre-trained weights) is used directly to encode the pairs. Its weights are fine-tuned in the end-to-end TDP training process.
\end{itemize}
%\vspace{-0.1cm}

All models use the same loss function and scoring as in \citet{zhang2018neural}. We present more details about the two BERT-based models below.

\vspace{-0.1cm}

\subsection{Model \textsc{bilstm-bert}} 

The first model adjusts the model architecture from \citet{zhang2018neural} to replace its word embeddings with frozen BERT embeddings. That is, word embeddings are computed via BERT for every sentence in the document; then, these word embeddings are processed as in the original model. More details about the BiLSTM model can be found in \citet{zhang2018neural}.

\vspace{-0.1cm}

\subsection{Model \textsc{bert-ft}} \label{section:model2}

This model takes advantage of BERT's multi-layer multi-head self-attention architecture \citep{vaswani2017attention} to learn feature representations for classification.
The embedding of the first token \verb|[CLS]| is interpreted as a classification output and fine-tuned. 

To represent a child-parent pair with context, \textsc{bert-ft} constructs a pseudo-sentence for the (potential) parent node and a pseudo-sentence for the child node. The pair of pseudo-sentences are concatenated and separated by the \verb|[SEP]| token, and then fed into the BERT model. Each pseudo-sentence is formed of the word(s) of the node, the node's label (TIMEX or EVENT), a separator token `:' and the sentence containing the node, as shown in Table \ref{table:2a_BERT_inputs}.

\begin{table}[H]
\centering
\begin{small}
\begin{tabular}{p{1.1cm}|p{1cm}|c|p{3.3cm}}
\hline
word(s) & label & sep & sentence \\
\hline
{\it February 27, 1998} & TIMEX & : & {\it Kuchma and Yeltsin signed a cooperation plan on February 27 1998.} \\
\hline
{\it called} & EVENT & : & {\it Yeltsin and Kuchma called for the ratification \ldots} \\
\hline
\end{tabular}
\end{small}
\caption{A pair of pseudo-sentences in \textsc{bert-ft}, for potential parent {\it February 27, 1998} and child {\it called} in Example 1 (The correct parent here is DCT).} \label{table:2a_BERT_inputs}
\vspace{-0.2cm}
\end{table}

\section{Experiments} 

We use the training, development and test datasets from \citet{zhang2019acquiring} for all experiments (182 train / 5 development / 9 test documents, total 2084 sentences). The documents in the datasets are already annotated with events and temporal expressions. This allows us to focus on evaluating the task of constructing temporal dependency trees.

We evaluated four configurations of the encoders above. Firstly \textsc{bilstm (re-implemented)} re-implements \citet{zhang2018neural}'s model\footnote{Originally implemented in DyNet~\citep{neubig2017dynet}.} in TensorFlow~\cite{abadi2016tensorflow} for fair comparison. Replacing its randomly-initialized embeddings with GloVe \citep{pennington2014glove} yields \textsc{bilstm-glove}. We also test the models \textsc{bilstm-bert} and \textsc{bert-ft} as described in Section \ref{section:approach}.

We used Adam \citep{kingma2014adam} as the optimizer and performed coarse-to-fine grid search for key parameters such as learning rate and number of epochs using the dev set~\footnote{We tried all parameter configurations with learning rates in $\{{0.001, 0.0001, 0.0005, 0.00025}\}$ and numbers of epochs in ${\{50, 75, 100\}}$, and perform 5 runs for each configuration. We observed a mean F1 of 0.58 with variance=0.002 across all configurations for all models. }. We observed that when fine-tuning BERT in the \textsc{bert-ft} model, a lower learning rate (0.0001) paired with more epochs (75) achieves top performance, compared to using learning rate 0.001 with 50 epochs for the BiLSTM models. We used NVIDIA Tesla P100 GPUs for training the models. On a single GPU, one epoch takes 7.5 minutes for the \textsc{bert-ft} model and 0.8 minutes for the \textsc{bilstm-bert} model.

\begin{table}[H]
\centering
\begin{small}
\begin{tabular}{p{5.4cm}|p{0.5cm}|p{0.5cm}}
\hline
\multicolumn{1}{l}{} & \multicolumn{2}{c}{F1 score} \\
Model & dev & test \\
\hline
Rule-based baseline~\citep{zhang2019acquiring} & 0.15 & 0.18 \\
BiLSTM~\citep{zhang2019acquiring} & 0.53 & 0.60 \\
\textsc{bilstm} (re-impl., \citealp{zhang2019acquiring}) & 0.45 & 0.55 \\  
\textsc{bilstm-glove} & 0.50 & 0.58 \\
\textsc{bilstm-bert} & 0.54 & 0.61 \\
\textsc{bert-ft} & 0.59 & {\bf0.68} \\
\hline
\end{tabular}
\end{small}
\caption{Performance of the models. }.
\label{table:model_scores}
\vspace{-0.6cm}
\end{table}

Table~\ref{table:model_scores} summarizes the F1 scores\footnote{Following ~\cite{zhang2019acquiring}, F1 scores are reported. For a document with $n$ nodes, the TDP task constructs a tree of $n+1$ edges, so F1 is essentially the same as the accuracy.} of our models. Results are averaged over 5 runs. We also include the rule-based baseline and the performance reported in \citet{zhang2019acquiring}, which applies the model of \citet{zhang2018neural} to the 2019 corpus, as a baseline\footnote{We were unable to replicate the F1-score reported in \citet{zhang2019acquiring} despite using similar hyperparameters. Therefore, we include performances for our re-implementation and the reported score in \citet{zhang2019acquiring} in Table~\ref{table:model_scores}.}. 

\textsc{bilstm-bert} outperforms the re-implemented \textsc{bilstm} model by 6 points and \textsc{bilstm-glove} by 3 points in F1-score, respectively. This indicates that the frozen, pre-trained BERT embeddings improve temporal relation extraction compared to either kind of non-contextualized embedding. Fine-tuning the BERT-based encoder (\textsc{bert-ft}) resulted in an absolute improvement of as much as 13 absolute F1 points over the BiLSTM re-implementation, and 8 F1 points over the reported results in \citet{zhang2019acquiring}. This demonstrates that contextualized word embeddings and the BERT architecture, pre-trained with large corpora and fine-tuned for this task, can significantly improve TDP. 

We also calculated the models' accuracies on time expressions or events subdivided by their type of parent: DCT, a time expression other than DCT, or another event. Difficult categories are children of DCT and children of events. We see that the main difference between \textsc{bilstm} and \textsc{bilstm-bert} is its performance on children of DCT: with BERT, it scores 0.48 instead of 0.38.
Conversely \textsc{bert-ft} sees improvements across the board over \textsc{bilstm}, with a 0.21 increase on children of DCT, a 0.14 increase for children of other time expressions, and a 0.11 increase for children of events.

\section{Analysis} 

{\bf Why BERT helps:} A detailed manual comparison of the dependency trees produced by the different models for articles in the test set shows BERT's advantages for TDP. The following phenomena are attested by many sentences in many documents and correspond to known properties of BERT.

Firstly, unlike \textsc{bilstm}, \textsc{bert-ft} is able to properly relate time expressions occurring syntactically after the event, such as {\it Kuchma and Yeltsin \underline{signed} a cooperation plan on \underline{February 27, 1998}} in Example 1. (\textsc{bilstm} falsely relates {\it signed} to the ``previous'' time expression DCT). This shows BERT's ability to ``look forward'' with its self-attention, attending to parents appearing after the child. 

Secondly, \textsc{bert-ft} is able to capture verb tense and use it to determine the correct relation for both DCT and chains of events. For example, it knows that present tense ({\it \underline{share} similar cultures}) overlaps DCT, while past perfect events ({\it was \underline{ruled} from Moscow}) happen either before DCT or before the event adjacent (salient) to them. 

Thirdly, \textsc{bert-ft} captures syntactic constructions with implicit temporal relations such as reported speech and gerunds (e.g. in Example 1, {\it Yeltsin and Kuchma \underline{called} for the ratification [\ldots], \underline{saying} it would \underline{create} \ldots}, it identifies that {\it called} and {\it saying} overlap and {\it create} is after {\it saying}). 

Similarly, BERT's ability to handle syntactic properties (\citealp{tenney2019bert}, \citealp{clark2019does}) such as embedded clauses
may allow it to detect the direction of connectives such as {\it since}.
While all models may identify the matrix clause verb as the correct parent, \textsc{bert-ft} is much more likely to choose the correct label. (\textsc{bilstm} almost always chooses `before' for DCT or `after' for children of events, ignoring the connective.)

Lastly, both \textsc{bert-ft} and \textsc{bilstm-bert} are much better than the \textsc{bilstm} at identifying context changes (new ``sections'') and linking these events to DCT rather than to a time expression in the previous sections  (evidenced by the scores on children of DCT). Because BERT's word embeddings use the sentence as context, the models using BERT may be able to ``compare'' the sentences and judge that they are unrelated despite being adjacent.

{\bf Equivalent TDP trees:} In cases where \textsc{bert-ft} is incorrect, it sometimes produces an equivalent or very similar tree (since relations such as \textit{overlap} are transitive, there may be multiple equivalent trees). Future work could involve developing a more flexible scoring function to account for this. 

{\bf Limitations:}  There are also limitations to \textsc{bert-ft}. For example, it is still fooled by syntactic ambiguity. Consider this example: 

{\bf Example 2:} {\it Foreign ministers agreed to set up a panel to investigate who shot down the Rwandan president's plane on April 6, 1994.}

A human reading this sentence will infer based on world knowledge that {\it April 6, 1994} should be attached to the embedded clause ({\it who shot down}), not to the matrix clause ({\it agreed}), but a syntactic parser would produce both parses. \textsc{bert-ft} incorrectly attaches {\it agreed} to {\it April 6, 1994}: even BERT's contextualized embeddings are not sufficient to identify the correct parse.

\section{Conclusion and Future Work}

We present two models that incorporate BERT into temporal dependency parsers, and observe significant gains compared to previous approaches. We present an analysis of where and how BERT helps with this challenging task.

For future research, we plan to explore other types of deep neural LMs such as Transformer-{XL}~\cite{dai-etal-2019-transformer} and XLNet~\cite{NIPS2019_8812}. 
As discussed in Section 5, we also plan to develop a more flexible scoring function which can handle equivalent trees.
Finally, we plan to evaluate \textsc{bert-ft} on other temporal relation datasets as part of a larger pipeline, which will include a mapping between TDTs and other temporal relation annotation schemas such as the TempEval-3 dataset \citep{uzzaman2013semeval}.

\section*{Acknowledgments}

This work was supported by DARPA/I2O and U.S. Air Force Research Laboratory Contract No. FA8650-17-C-7716 under the Causal Exploration program, and DARPA/I2O and U.S. Army Research Office Contract No. W911NF-18-C-0003 under the World Modelers program. The views, opinions, and/or findings expressed are those of the author(s) and should not be interpreted as representing the official views or policies of the Department of Defense or the U.S. Government. This document does not contain technology or technical data controlled under either the U.S. International Traffic in Arms Regulations or the U.S. Export Administration Regulations.

\bibliography{emnlp2020}
\bibliographystyle{acl_natbib}

\end{document}